\documentclass[conference]{IEEEtran}
\usepackage{times}

\usepackage[numbers]{natbib}
\usepackage{multicol}
\usepackage[bookmarks=true]{hyperref}

\usepackage{subfigure}
\usepackage{microtype}
\usepackage{graphicx}
\usepackage{booktabs} 
\usepackage{hyperref}

\begin{document}

\title{Closing the Accuracy Gap in an Event-Based Visual Recognition Task}

%


%
\author{\authorblockN{Bodo R\"uckauer\authorrefmark{1},
Nicolas K\"anzig\authorrefmark{2},
Shih-Chii Liu\authorrefmark{1}, 
Tobi Delbruck\authorrefmark{1} and
Yulia Sandamirskaya\authorrefmark{1}}
\authorblockA{\authorrefmark{1}Institute of Neuroinformatics, University of Zurich and ETH Zurich, Switzerland \\
Email: ysandamirskaya@ini.uzh.ch}
\authorblockA{\authorrefmark{2}Department of Information Technology and Electrical Engineering, ETH Zurich, Switzerland}}

\maketitle

\begin{abstract}
Mobile and embedded applications require neural networks-based pattern recognition systems to perform well under a tight computational budget. In contrast to commonly used synchronous, frame-based vision systems and CNNs, asynchronous, spiking neural networks driven by event-based visual input respond with low latency to sparse, salient features in the input, leading to high efficiency at run-time. The discrete nature of the event-based data streams makes direct training of asynchronous neural networks challenging. This paper studies asynchronous spiking neural networks, obtained by conversion from a conventional CNN trained on frame-based data. As an example, we consider a CNN trained to steer a robot to follow a moving target. We identify possible pitfalls of the conversion and demonstrate how the proposed solutions bring the classification accuracy of the asynchronous network to only 3\% below the performance of the original synchronous CNN, while requiring 12x fewer computations. While being applied to a simple task, this work is an important step towards low-power, fast, and embedded neural networks-based vision solutions for robotic applications.
\end{abstract}

\IEEEpeerreviewmaketitle

\section{Introduction}
Deep convolutional neural networks (CNNs) offer practical solutions for pattern recognition and have radically changed the field of image recognition. In the field of robotics, however, where real-time processing and low power budget are crucial, CNN-based image-processing algorithms face a fundamental latency-power trade-off, where low latency can only be achieved by dramatically increasing power consumption.

Evolution of embedded systems led to development of event-based vision, which has enabled improved performance of vision systems for fast and agile robots~\cite{FalangaEtAl2017,Mueggler2017}. Event-driven, biologically inspired vision sensors such as DVS~\cite{lichtsteiner2008128}, ATIS~\cite{posch2011qvga}, and DAVIS~\cite{brandli2014240} enable fast and low-power processing of visual information. Instead of capturing static images of the scene, these sensors record pixel brightness change events with high temporal precision. Events are only triggered if a significant change occurs in the observed scene, allowing lower latency and lower required bandwidth compared to frame-based sensors. However, since the data produced by an event sensor is a sequence of events, conventional frame-based computer-vision algorithms~\cite{opencv_library} or DNN-based pattern recognition can not be applied directly.   

One obvious way to process event-based data with conventional DNNs is to create frames by accumulating events over fixed time intervals, or accumulating a constant number of events per frame. It has been shown in several robotic applications that by following this approach, conventional CNNs can be applied for feature extraction and object classification \cite{moeys2016steering,amirlow,lungu2017live}. Although using constant event count frames addresses the latency-power tradeoff by using data driven computation, it ignores key advantages of event based sensors, in particular their sparse data and the high temporal precision.

This paper explores the use of asynchronous neural network architectures for processing the event-based vision data. In contrast to the synchronous, frame-based mode of operation of conventional CNNs, asynchronous spiking neural network (SNN) architectures 
represent hidden layer activations in form of discrete events -- spikes -- that are propagated through the network asynchronously, so that neurons are only activated when they receive events~\cite{martin2015spiking}. Theory has shown that SNNs are at least as computationally powerful as conventional neuronal models being used in deep-learning \cite{dada}. It has also been shown that by the use of dedicated event-based hardware, power consumption and latency can be reduced by several orders of magnitude \cite{furber2013overview,IndiveriCorradiQiao2015, merolla2014million}. IBM's TrueNorth processor \cite{merolla2014million} consumes about 1000 times less energy than conventional synchronous architectures. Thus, just as hardware acceleration through GPUs has played a fundamental role in the advancements of deep-learning, there is increasing availability of neuromorphic SNN accelerators that enable efficient computation of event-based SNN training and inference \cite{lee2016training}, potentially running on a fraction of the energy budget compared to conventional CNNs running on GPUs~\cite{furber2013overview,Qiao2015,IndiveriChiccaDouglas2009,merolla2014million}.

There are two ways to obtain an SNN for solving a pattern recognition task. First, recent work has explored direct training in the spiking domain using backpropagation inspired techniques for training multi-layer SNN architectures \cite{lee2016training,neftci2016neuromorphic,neftci2017event}. Training SNNs is difficult as due to their non-differentiable nature and gradient-descent based methods can not be applied directly. Furthermore, backpropagation rules typically used in deep learning rely on the availability of network-wide information stored with high-precision memory, and on precise operations that are difficult to realize in event-based hardware~\cite{neftci2017event}. 

Second, SNNs can be constructed by converting conventionally trained analog neural networks (ANNs)~\cite{Rueckauer2017Conversion,diehl2016conversion}. In terms of accuracy, \cite{Stromatias2017} reports that while these neural networks seem to work well using synthetic input spike trains generated artificially from frame images (e.g., from the MNIST database), where the gray level of an image pixel is transformed into a stream of spikes, doing inference using this SNN with data from an event-based vision sensor may lead to significant loss in accuracy. Increasing our understanding of SNN processing is needed to close the accuracy gap between the frame-based and event-based pattern recognition.

In this paper, we apply the second method and analyze object recognition using analog and spiking convolutional neural networks in the context of a robotics predator/prey navigation scenario. The dataset from \cite{moeys2016steering} is used to train and evaluate several neural network architectures, where the purpose of the trained networks is to steer a predator robot in the direction of a prey robot. In particular, we compare the conventional CNN architecture proposed in \cite{moeys2016steering} with its event-based SNN counterpart, where accuracy is evaluated using both synthetic and sensor-driven input spike trains. We perform a thorough analysis of accuracy losses that occur in the ANN to SNN conversion and offer solutions to reduce these losses.  We show that a CNN trained on constant event count frames can be run efficiently on the asynchronous sensor events at inference, using up to 12x fewer computations than when using frames. Finally, we identify the causes for classification accuracy loss that occurs when switching from a synchronous training mode to an asynchronous inference mode, and evaluate solutions that minimize this loss.

These are crucial steps on the way to low-power, fast, embedded solutions, which will enable the application of deep neural networks on robotic platforms in real time and with a limited power budget. 


\section{Methods}

This section describes the pipeline of the present work: Starting with an event-based data set (Sec.~\ref{sec:data}), we first synthesized frames (Sec.~\ref{sec:frame_generation}) to train a conventional ANN (Sec.~\ref{sec:training}). The resulting frame-based model was then converted to an SNN (Sec. \ref{sec:conversion}) and tested on the original event-based data (Sec.~\ref{sec:testing}).

\subsection{The data set}
\label{sec:data}

The data set from \cite{moeys2016steering} consists of twenty recordings with a total duration of 1.25 hours from a Dynamic and Active Pixel Sensor (DAVIS)~\cite{brandli2014240}. The DAVIS camera was mounted on the predator robot and recorded different scenes in which the predator robot, driven by a human or by the CNN, followed the prey robot. The recordings contain both conventional image frames (APS) as well as event-based data (DVS). The APS frames were not used in this work. The DVS sensor data was output in AEDAT 2.0 file format~\cite{aedat}: each sensor event contained a timestamp, the pixel address, and a polarity value (ON/OFF), indicating an increase or decrease in pixel brightness.

The ground truth labels of the recordings encoded the position and bounding box of the prey robot. From these labels, we produced a target ground truth of four classes, marking in which third of the visual field the prey robot is located (classes 1-3) or if it is not visible (class 4), leading to a four-class-classification problem (left, center, right, invisible). The DAVIS camera records with a resolution of 240x180 pixels. We subsampled the event-addresses to 36x36 arrays. \cite{moeys2016steering} found this to be the minimum size for which the robot can still be recognized by human eye. The data set consisted of roughly 200k images generated by binning DVS events to 5000-event frames, as described below.

\begin{figure}[t!] 
\centering
\subfigure[\label{fig:fig_3a}]{
	\includegraphics[height=0.2\linewidth]{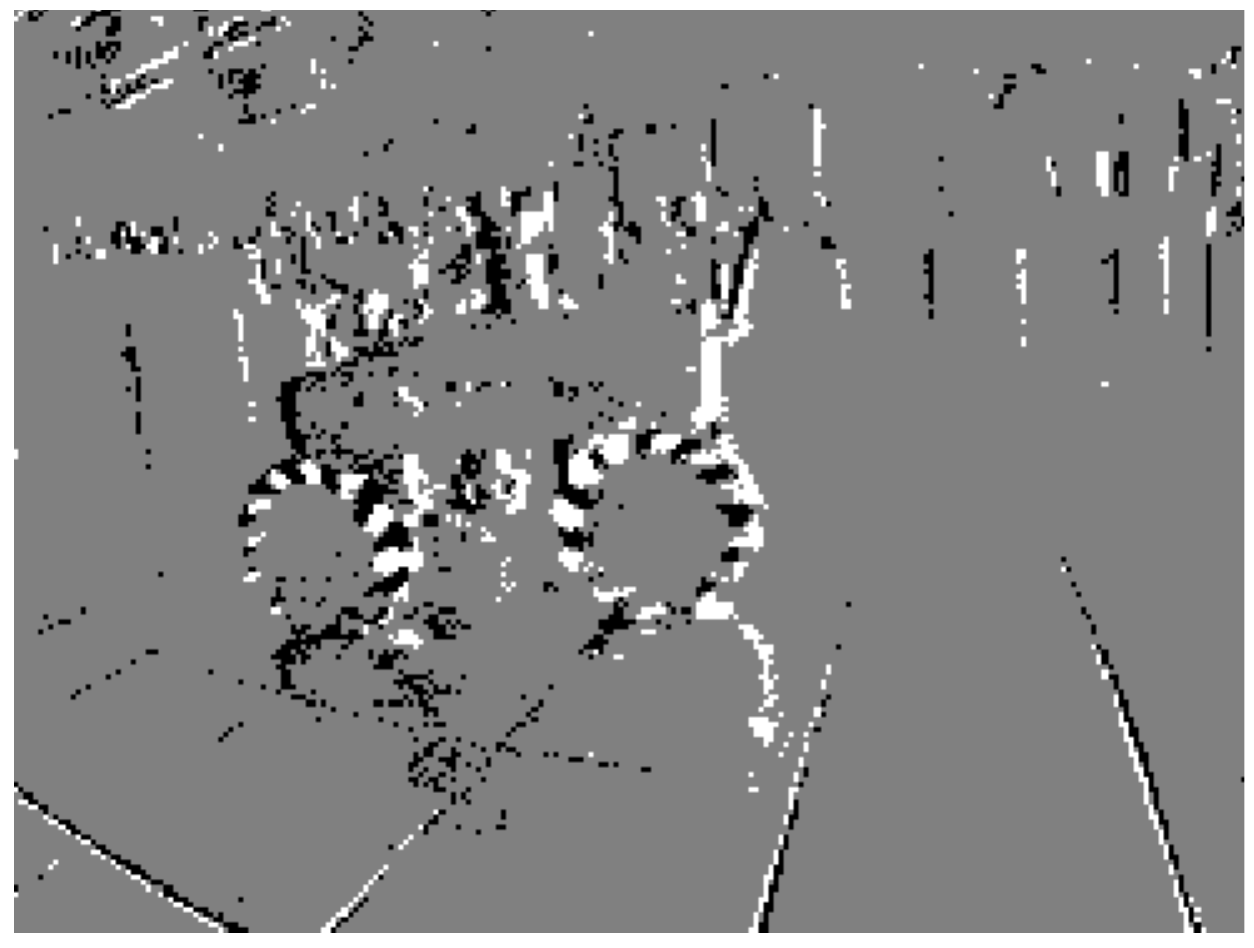}
    \includegraphics[height=0.2\linewidth]{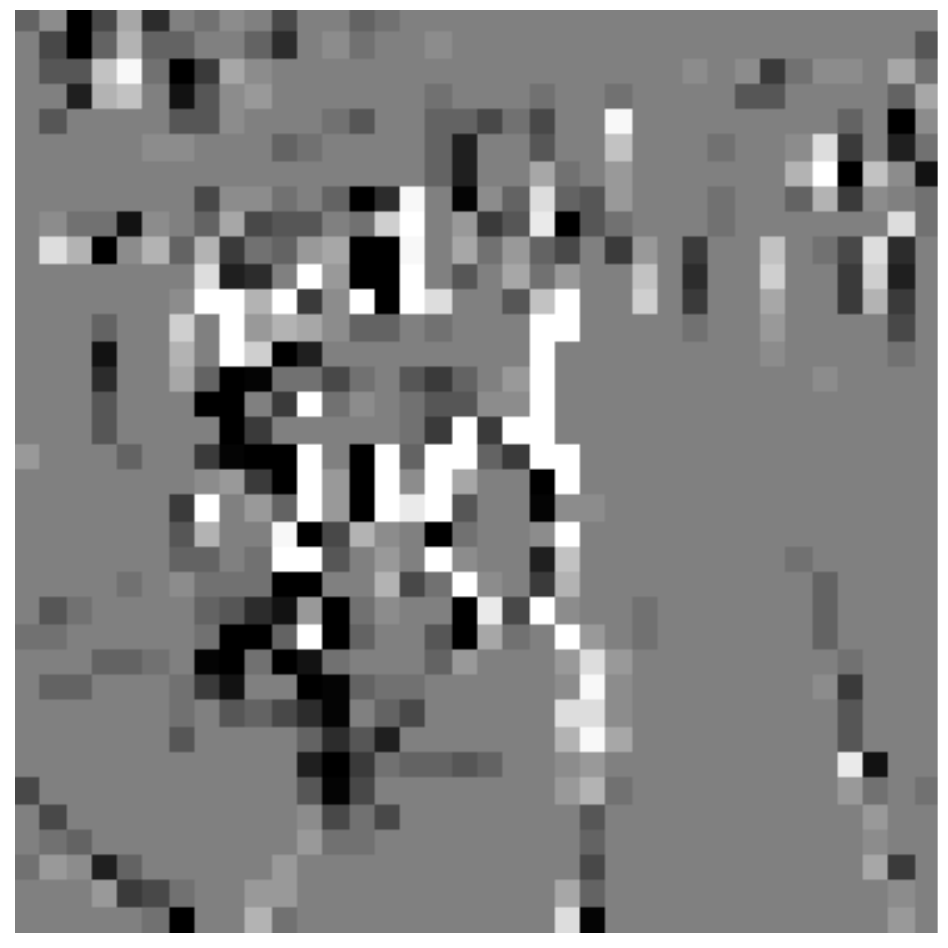}}	
\hfill
\subfigure[\label{fig:fig_3b}]{
	\includegraphics[height=0.2\linewidth]{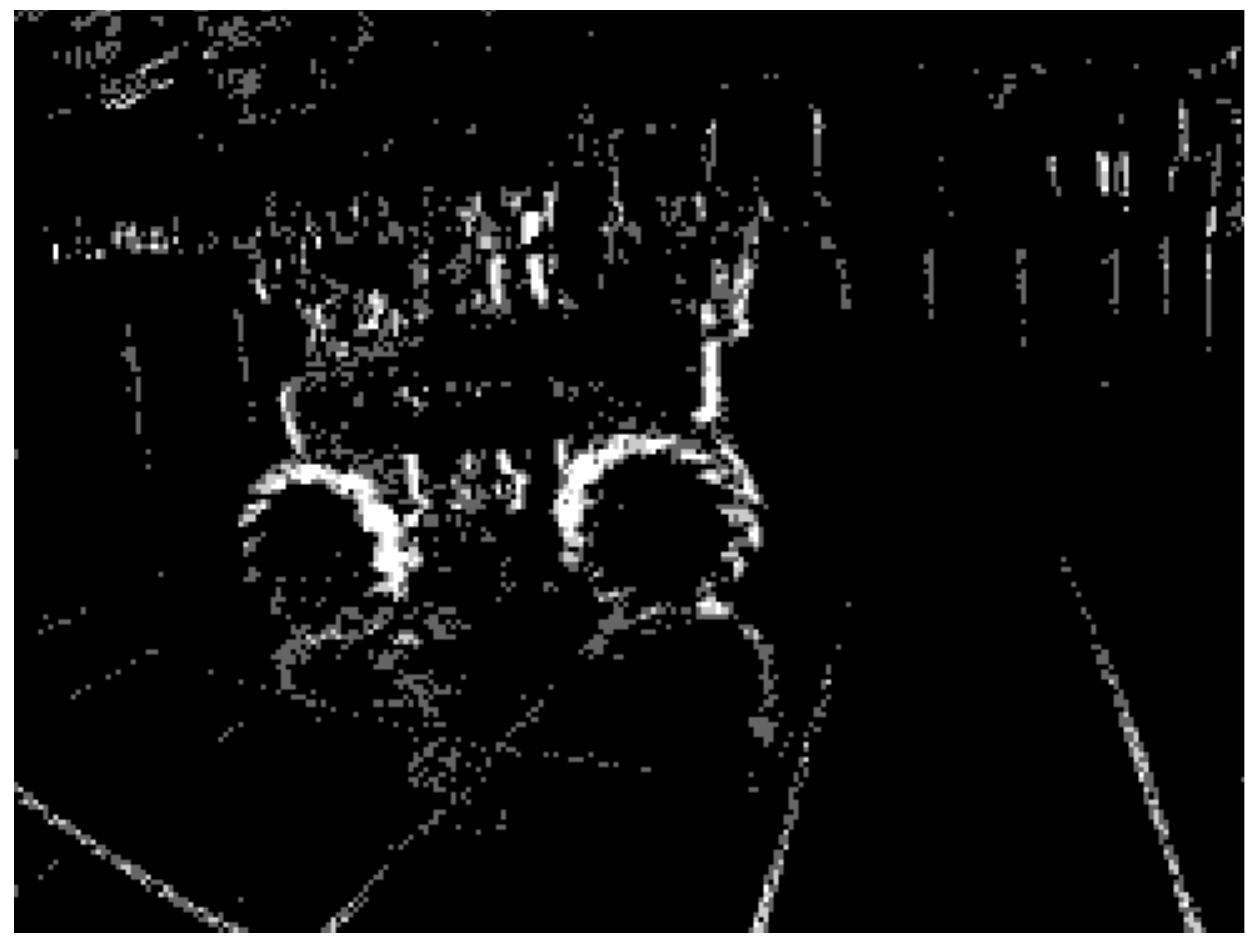}
	\includegraphics[height=0.2\linewidth]{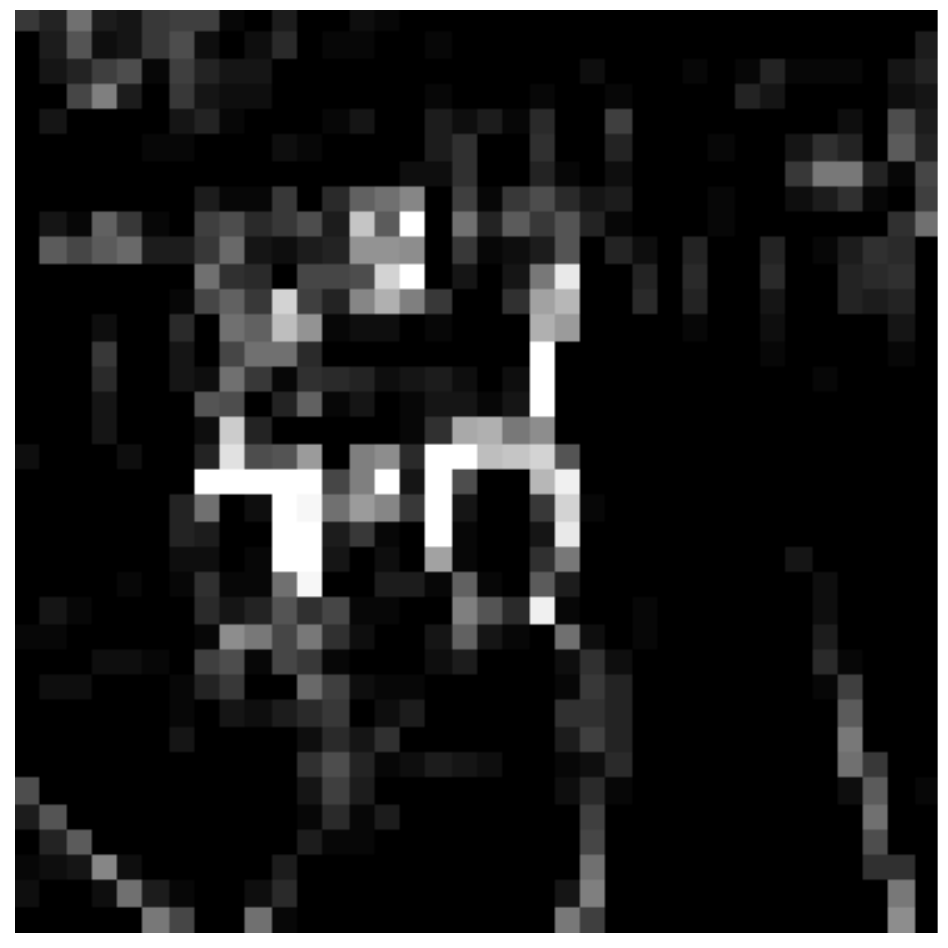}}
\caption{Synthesized DVS frames (see Sec. \ref{sec:frame_generation} for details). \ref{fig:fig_3a}: Original method from \cite{moeys2016steering}. \ref{fig:fig_3b}: Our method. Left column: Full resolution (240x180); right column: Subsampled to 36x36.}
\label{fig:frames}
\end{figure}

\subsection{Generating frames from event-based data for ANN training}
\label{sec:frame_generation}

Training of the ANN was conducted on frames synthesized from DVS events, but testing of the converted SNN was performed on the original DVS stream. Thus, any transformation of the data during frame synthesis should either be applicable to the underlying DVS event streams as well, or else distort them as little as possible. This section describes each step of a frame generation method that best preserves the classification performance when using asynchronous DVS data at test time.

\subsubsection{Choosing the binning window}
Frames can be synthesized from DVS data either by accumulating a variable number of events during a fixed time window, or by accumulating a fixed number of events during a variable time window. We follow \cite{moeys2016steering} and use the latter approach, with a constant number of 5000 events per frame. This way, the frame rate is proportional to the rate of change of the scene, and each frame is more likely to be informative. Frame synthesis with a fixed time window can lead to overly sparse and noisy frames during time intervals with few changes in the recorded scene, and blurred frames when the robots are moving quickly.

\subsubsection{Handling polarity}
Another design choice concerns the polarity of the DVS events. One can integrate ON and OFF events by representing them as +1 or -1, respectively. In this case, events of opposite polarity cancel each other. The original work \cite{moeys2016steering} uses this method by initializing the frames with 0.5 pixel intensity and in-/decrementing this value by $\pm 0.005$ per ON/OFF event. This approach is not feasible in our setup, because it assigns a nonzero intensity to pixels where the DVS records no events. Instead, we start with an all-zero frame and apply a rectified event count that discards polarity while binning the events. In preliminary experiments we found the polarity information not to be relevant for learning this task.

\subsubsection{Input normalization}
Outliers in the distribution of event counts were removed by clipping values greater than three times the standard deviation (3-sigma normalization).

Though the network was trained on frames, we aimed to use the original DVS events during inference. To maintain high classification accuracy when switching from frame to event input, any transformation of the frame data, performed during training, should be applied to the DVS data as well. 3-sigma normalization on DVS event streams is possible by temporally binning 5000 events into a frame as during training, and applying 3-sigma normalization on this frame of integer event-counts to identify outlier events, which are then removed from the DVS event stream fed into the SNN. 


\subsubsection{Input scaling}
After accumulation and outlier removal, the frames are scaled to $[0, 1]$ real values, which is the last stage of synthesizing frames for training the ANN. During testing of the SNN, discrete events are streamed into the network, making scaling inapplicable. 

A subtle difference in the training-frames arises from the order in which scaling and 3-sigma normalization are applied. In \cite{moeys2016steering}, the frame of integer event-counts is scaled to $[0, 1]$ before 3-sigma normalization. The resulting frames consist of real values. If instead 3-sigma normalization is applied first (by removing discrete events from the frame of integer event-counts), the subsequent scaling will result in frames that consist of rational numbers only.
To avoid the discrepancy of training on real values and testing on integers, we applied 3-sigma normalization \textit{before} scaling. This seemingly small difference turns out to be crucial: With the opposite ordering, the classification accuracy of the converted SNN drops by 30\%.

Fig.~\ref{fig:frames} shows an example of the frames synthesized as described above. In Table \ref{tab:methodology} 
we summarize how our approach differs from the original work \cite{moeys2016steering}.

\begin{figure}[t!]
  \centering
  \includegraphics[width=1.0\linewidth]{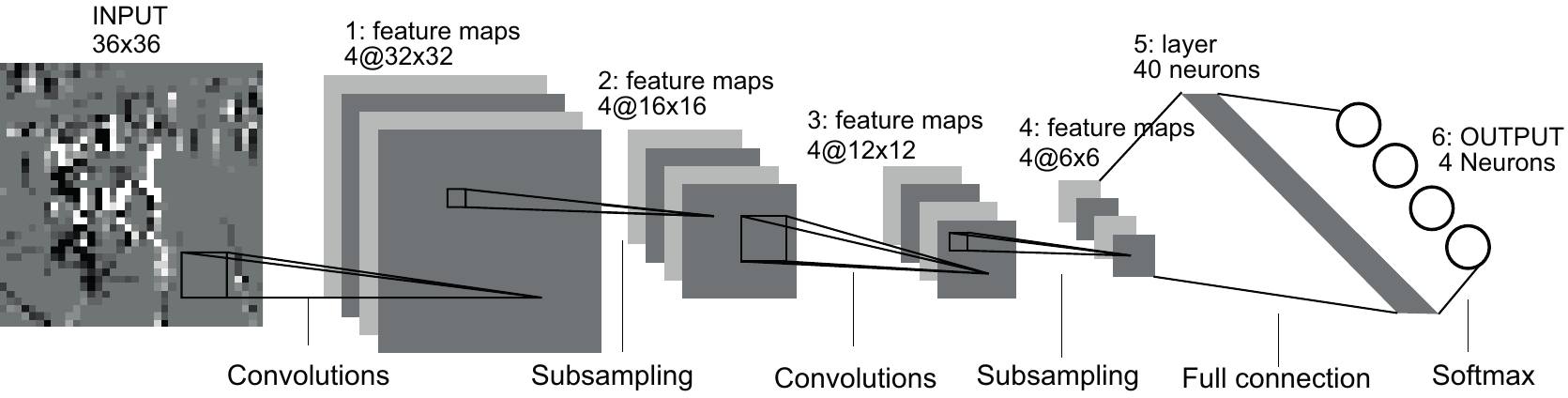}
  \caption{Architecture for predator-prey task. (Graphic adapted from \cite{lecun1998gradient})}
  \label{fig:cnn}
\end{figure}
 
\subsection{Training the frame-based ANN}
\label{sec:training}

Fig. \ref{fig:cnn} shows the model architecture that \cite{moeys2016steering} developed to solve the predator-prey task. It consists of a small CNN with two convolution layers with 4 feature maps each and a kernel size of 5x5 pixels. Each convolution layer is followed by a max pooling layer. A fully-connected layer of 40 neurons connects the last pooling layer with the 4 classifier output units. The network contains a total of 5884 neurons and 6472 parameters. \cite{moeys2016steering} showed that this tiny CNN achieves higher classification accuracy than humans observing the same images.

We implemented the network in the Keras framework~\cite{chollet2015keras}, and trained for 30 epochs using mini-batches of size 32 and the ADAM optimizer.

\subsection{Converting the ANN to an event-driven SNN}
\label{sec:conversion}

The central idea of ANN-SNN conversion is that the time-averaged firing-rates of the 
resulting spiking architecture correspond to the analog activations in the original ANN. 
This mapping can be achieved by replacing the neurons in the ANN 
with non-leaky integrate-and-fire neurons~\cite{burkitt2006review}. 
The trained parameters remain the same, 
up to a layer-wise rescaling that reduces the problem of limited dynamic range of spiking neurons~\cite{diehl2015fast}. 
\cite{Rueckauer2017Conversion} proposed implementations in the spike domain of 
modules commonly used in ANNs, 
like max-pooling and softmax layers.
We apply their open-source conversion framework~\cite{bodotoolbox} 
to transfer the predator-prey ANN to the event-based domain.

\begin{table}[t!]
\caption{Comparison of methodology.}
\centering
\resizebox{0.47\textwidth}{!}{%
\begin{tabular}{ l l l}
& \cite{moeys2016steering} & This work \\ 
\hline
\vspace{-0.2cm} \\
APS frames used & Yes & No \\
Biases & Yes & Yes, with L2-regularizer \\
Event polarity used & Yes & Yes, rectified \\
Frames initialized at & 0.5 & 0,$\;$ for comput. sparsity \\
Subsampling (cf. \ref{sec:subsampling}) & $\mathrm{sum}$ & $\mathrm{max}$,$\;$ to remove clusters \\
Scaling frames to $[0, 1]$ & Yes & Yes (N/A in SNN) \\
3-sigma normalization & Yes & Yes, before scaling \\
\end{tabular}}
\label{tab:methodology}
\end{table}

\subsection{Data used for testing the converted SNN}
\label{sec:testing}
Due to the lack of truly event-based datasets acquired with neuromorphic vision sensors, in recent work the spiketrains were often generated synthetically from frame-based image datasets. The most common method is to use poisson spike generators driving their firing-rate with the intensity of the corresponding input pixels. However, the stochastic nature of the generated Poisson spike trains introduces noise into the network, without having any notable benefits. A simple alternative is to use analog input values in the very first hidden layer, and to compute with spikes from there on~\cite{zambrano2016fast,Rueckauer2017Conversion}. The image pixel values are interpreted as currents flowing into the neurons of the first hidden layer, where they are integrated into membrane potentials, thus deterministically producing regular spikes at a rate proportional to the pixel value. Recent work \cite{Stromatias2017} has reported that while converted SNNs seem to work well on synthetic input data, using real event-based data as input can lead to a significant drop in classification accuracy.

In this work, we perform simulations with \textit{poisson} and \textit{analog} inputs from synthetically generated frames (Sec. \ref{sec:frame_generation}), and we compare these to directly applying the original DVS events from the predator/prey dataset as input spikes.

\subsection{Simulation of the converted SNN}
We make use of the SNN toolbox \cite{bodotoolbox} to run the converted SNN on the three input types described above. The SNN toolbox provides a simulator for spiking networks that is built on the Keras framework. The spiking network consisting of non-leaky integrate-and-fire neurons is processed in a time-stepped manner with a step size equivalent to the time resolution of the DVS event stream (1 microsecond). 

The DVS data set used for testing the converted SNN is stored as a collection of \textit{.aedat} files, where each file contains a DVS clip of several seconds. Previously, the toolbox accepted frame-like input. To be able to use asynchronous data, we extended the toolbox by a DataGenerator module that iteratively reads in an .aedat file and processes the event sequence with subsampling and outlier removal as described in Sec. \ref{sec:frame_generation}. The network outputs a classification guess at each time step, but we define the period of time needed to process 5000 events as ``one sample'', and take as final classification output of the network for one particular sample the class corresponding to the neuron that fired the most spikes while processing the sample. When all events in an .aedat file are processed, the DataGenerator loads the next sequence of events from the aedat-directory; this procedure is repeated until all events are processed.

\section{Results}
\label{sec:results}

\subsection{ANN accuracy}
\label{sec:ann_results}
First, we reproduced the results of \cite{moeys2016steering} 
by training the frame-based CNN architecture in Fig. \ref{fig:cnn} on frames generated from the DVS events as outlined in Sec.~\ref{sec:frame_generation}. 
The original work used a combined dataset of APS frames and 
synthesized DVS frames. 
We found that the same classification performance can be achieved using DVS frames alone, 
which is preferable in the present setup, because only DVS events will be used during inference.

\begin{figure*}[!ht]
\centering
\includegraphics[width=1.0\linewidth]{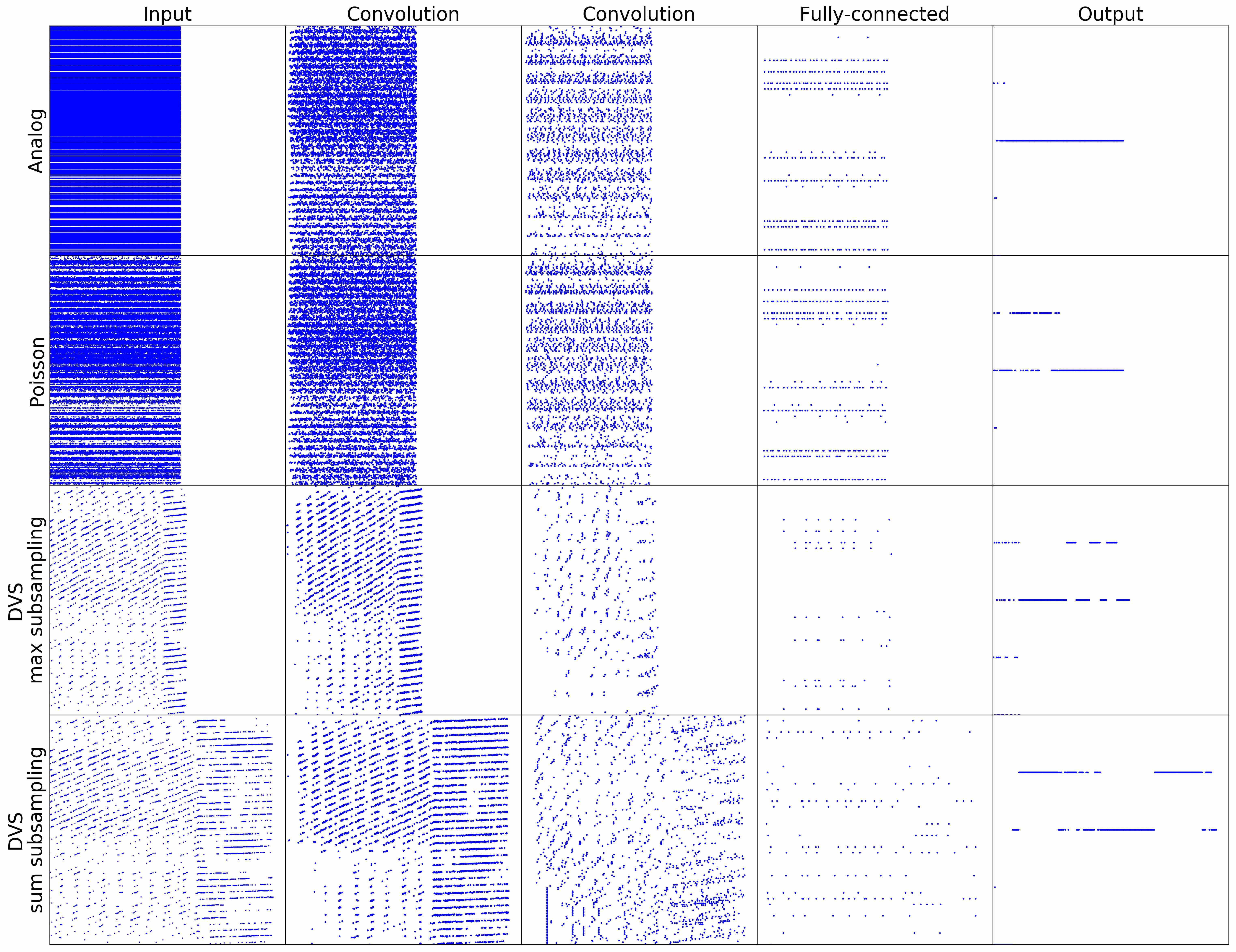}
\caption{Spike trains generated by simulating the SNN on a single test sample corresponding to 5000 DVS events. X-axis: time (450 steps of simulation); y-axis: neuron index. See Sec. \ref{sec:testing} for a description of the input types, and Sec. \ref{sec:frame_generation} for details on the frame generation. The Figure is discussed at the end of Sec.~\ref{sec:dvs_spiketrains}.}
\label{fig:spiketrains}
\end{figure*}

\subsection{SNN accuracy}

After transferring the ANN to an SNN as described in 
Sec.~\ref{sec:conversion}, the SNN was tested on the three input types 
listed in Sec.~\ref{sec:testing}, namely 
\textit{analog} (frame-based), 
\textit{Poisson}, and \textit{DVS} input. 
Both analog and Poisson input resulted in SNN accuracies 
close to the original ANN accuracy (see Table~\ref{tab:accuracy}).
However, in our initial experiments, 
the SNN accuracy dropped to chance level when using the original DVS input spike trains. 
We discuss the reasons for this reduction of accuracy, 
and propose and evaluate solutions to restore accuracy in the remainder of this section.

\subsubsection{Imbalance between network biases and DVS rates}
In \cite{bodotoolbox}, the bias values of neurons in the 
ANN are converted into constant input currents that flow into SNN 
neurons over the course of the simulation. 
If a bias value is large, this bias current can outweigh the spike-driven input to a neuron and dominate that neuron's output firing dynamics. 
This effect is likely to occur in neurons receiving DVS input spike trains, 
whose rates may vary considerably over the duration of a single 5000-event sequence (see Sec.~\ref{sec:dvs_spiketrains}). 

To prevent dominating biases, we trained the ANN with L2-regularization on the network weights and biases. L2-regularization adds to the training cost function a term which is proportional to the squared parameter values, thereby inducing the network to keep parameter values small. Training the ANN without regularization led to several bias values that were close to or above the threshold of the SNN neurons, thereby dominating their firing dynamics. The classification accuracy of the converted L2-regularized SNN 
increased by 43\% as compared to the SNN without regularization. 

Training the ANN altogether without biases (as done in \cite{diehl2016conversion}) may be another straight-forward solution. We trained an ANN without biases, which achieved 0.6\% lower accuracy than the L2-regularized ANN containing biases. The converted SNN without biases scored better than the non-regularized SNN with biases, but 0.81\% worse than the L2-regularized SNN with biases. Thus, we favor the regularized model with biases.

\subsubsection{Subsampling induces temporal clusters}\label{sec:subsampling}
Aside from dominating biases, another reason for the drop in classification performance when using DVS input was the subsampling mechanism. Pixel addresses in the original 240x180 image space are subsampled to 36x36 by integer division. If a subsampled patch contains several simultaneous events, they will all be mapped onto a single pixel address, thereby transforming a spatial-temporal cluster into a temporal cluster. A neuron in the SNN then receives the spikes contained in such a ``burst'' at immediately-subsequent time steps during the simulation. These spike bursts are in strong contrast to the way the network was trained, namely on analog frames, where such temporal structures are not present.

To see why temporally structured spike trains may produce a different outcome than homogeneously distributed spike trains, consider a neuron receiving a fixed number of input spikes from two sources, one inhibitory and the other one excitatory. If both sources fire at a regular rate, their contributions cancel each other and the neuron will not be active. If instead the spikes of the excitatory source are clustered into an early spike burst, the neuron will be strongly activated, even though the total number of spikes from each source over a given time period has not changed.

To prevent formation of detrimental temporal clusters during subsampling, we keep only one of the subsampled events in a patch. Here we term this method \textit{max subsampling}, and the method that accumulates all events in a patch \textit{sum subsampling}. The classification accuracy of the ANN trained on the max- and sum-subsampled data is 88.25\% and 88.04\%, respectively. The accuracy of the converted SNN is 85.19\% and 78.24\%, respectively, which shows the importance of removing spike bursts due to subsampling.

\subsubsection{Non-uniform DVS spiketrains} \label{sec:dvs_spiketrains}
With regularized biases and max-subsampling, the classification accuracy of the converted SNN using DVS input improves from chance level to within 3\% of the original ANN. We were not able to close this gap completely, and believe the underlying cause to be inhomogeneities in the DVS spike trains.

By viewing the DVS recordings as well as by studying the raster plots in Fig.~\ref{fig:spiketrains}, one can observe phases of 
increased global activity within the time window that corresponds to 5000 events. 
These bursts of global activity are likely the result of electrical coupling between frame electronic shutter and DVS circuits within the pixels~\cite{brandli2014240}. 
These abrupt changes in firing frequency propagate throughout the network. 
The variability of DVS event rates differs strongly from the rather uniform spike distribution observed when using Poisson or analog input. 
In the previous subsection, we argued that temporal spike patterns 
in the test phase can have a detrimental effect because of the 
asymmetric spike generation mechanism: 
neuron activity due to a burst of excitatory input cannot be reversed by a 
later burst of inhibitory input. 
We can not expect the network to be able to cope with temporal structure in the input which it has never experienced during training.

This intuitive explanation of the remaining accuracy loss can be validated by using Poisson or analog input, which features constant firing rates as during training. 
With 88.12\% accuracy, analog input nearly closes the accuracy gap.
With 86.77\% accuracy, Poisson input falls between analog and DVS input,
which is reasonable given the variance inherent in a Poisson process.

Figure \ref{fig:spiketrains} compares the spike trains generated by the different input types. \textbf{Analog input} (first row) results in the most regular firing dynamics and the highest support for the correct class label (second neuron in output layer). Surprisingly, the slight variations induced by \textbf{Poisson} variability (second row) reduce the network's confidence in the correct class label significantly. Asynchronous \textbf{DVS input} (third row) exhibits temporal structure that is not present in analog or Poisson input. The spike rates are generally lower, which accounts for the reduced operation cost, but contributes also to the increased classification error. DVS input with sum-subsampling (bottom row, cf. Sec.~\ref{sec:subsampling}) contains spike bursts that are fed into the network in close succession, causing the spike train to spread out over a longer simulation time. These temporal patterns -- unseen during training -- cause the network to confuse the correct output label (right column).

\begin{figure}[t!]
  \centering
  \includegraphics[width=0.95\linewidth]{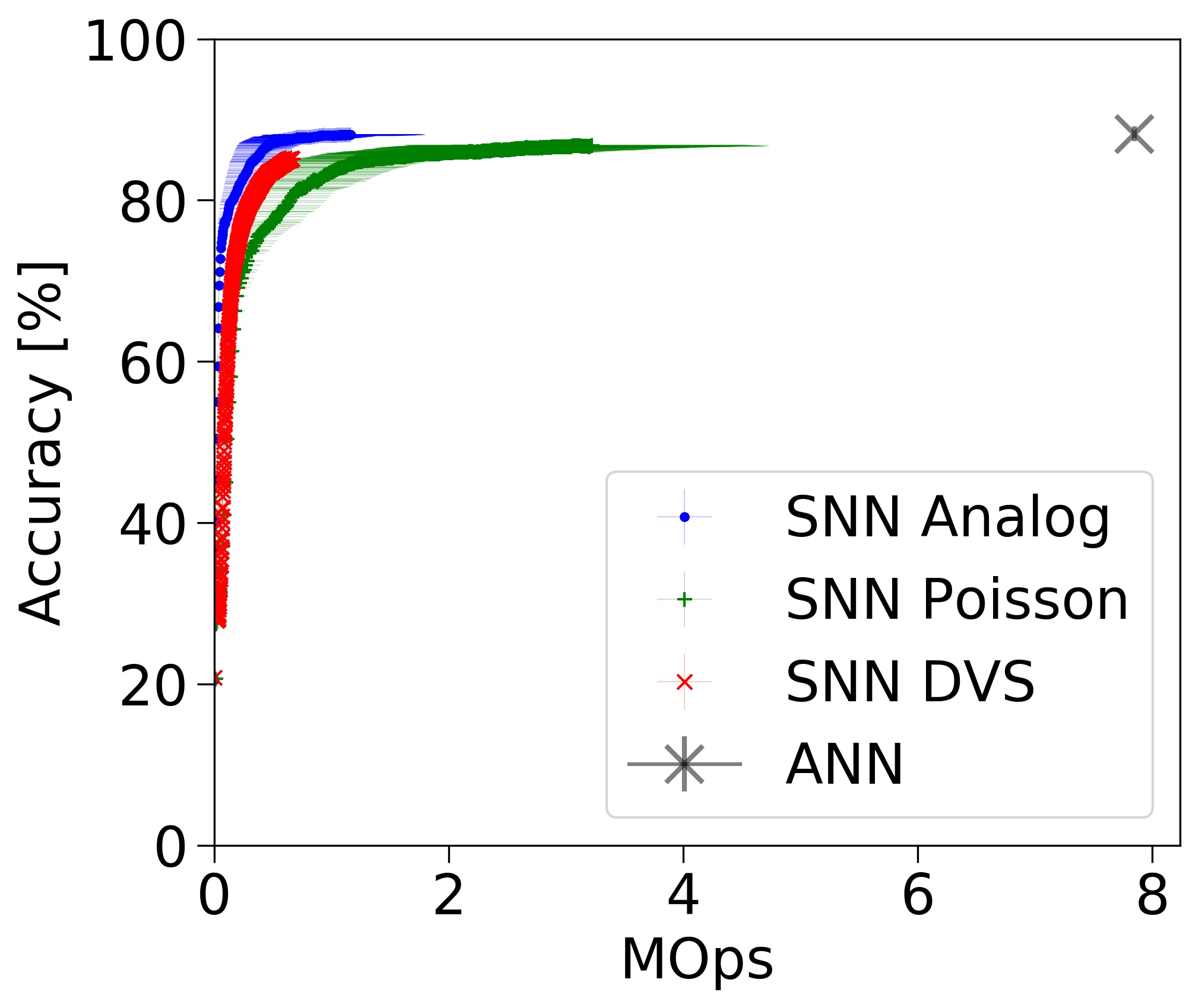}
  \caption{Average classification test accuracy of ANN (single cross) and SNNs (curves) for DVS, analog and Poisson input, plotted against the number of operations.}
  \label{fig:err_vs_ops}
\end{figure}

\subsection{Operation cost}
Besides classification accuracy, a second important metric for SNN performance is its operation cost. 
An ``operation'' for the SNN is defined as a synaptic update, i.e., the update of a neuron's state due to a spike in the preceding layer. This operation corresponds to a simple ``addition'', in contrast to more costly multiply-accumulate operations needed in conventional ANNs. We compute this quantity from the network architecture and the number of spikes that each neuron fires during the simulation~\cite{Rueckauer2017Conversion}. 

Figure \ref{fig:err_vs_ops} compares the accuracies and operation costs of the ANN trained with L2-regularized biases on max-subsampled data, and of the converted SNN tested on the original DVS events, synthesized analog frames, and Poisson spike trains. The operation cost for the ANN is a single value, because inference consists of a single forward pass. In the SNN, a continuum of classification accuracies is obtained as simulation progresses and more operations are invested. Table~\ref{tab:accuracy} lists the final accuracy and operation cost at the end of each of the SNN curves. 

Summarizing the results, the SNN with analog input provides the highest accuracy while reducing the number of operations by 7x compared to the original ANN. The SNN with DVS input suffers from an accuracy loss of 3\%, but compensates for it by a 12x reduction in computational cost.

\section{Conclusion} 
\label{sec:conclusion}

In this work, we explored spiking neural networks (SNNs) as efficient replacements of conventional frame-based, analog neural networks (ANNs), on the task of a robot pursuing a moving target. The underlying data set stems from a dynamic vision sensor (DVS), which provides a continuous stream of asynchronous events. These event streams are seldomly used directly as input to deep neural networks; instead, the events are commonly binned into frames, on which the network is trained and tested. While this frame-based approach grants easy access to a wealth of powerful deep learning frameworks, one sacrifices the advantage of very low latency and potentially sparse computation inherent in asynchronous event streams from a DVS. Converting a pre-trained frame-based ANN into an event-driven SNN aims to combine the best of both worlds: Frame-based training provides us with a high-accuracy model, while inference is done on sparse asynchronous events.

This study confirms earlier findings \cite{Rueckauer2017Conversion,zambrano2016fast} showing that the converted SNN achieves equivalent classification accuracy as the original ANN when using static frames or Poisson spike trains as input. However, we take this analysis a step further by applying the original DVS events as input to the SNN. Initial classification results were close to chance level, indicating significant distortions when transitioning from a synchronously trained model to an asynchronously tested model. As causes for this accuracy loss we identify (1) the way that the training frames are generated from DVS data, (2) extreme weights or biases, and (3) temporal structure present in the asynchronous test data but not in the training frames. 

To solve these issues, we propose (1) training frame generation steps that are applicable to the DVS test data as well, thereby minimizing the discrepancy between training and test set, and (2) L2-regularization during training to effectively prevent dominating model parameters. The resulting SNN achieves classification accuracy close to the original ANN. The third issue, temporal structure in the DVS event stream, can only partly be removed, and likely accounts for the remaining 3\% accuracy gap between synchronous ANN and event-driven SNN. 

By evaluating the computational cost of the SNN when run on DVS events, we confirm the expected improvement in terms of low latency and sparse, change-driven operation. Specifically, inference in the SNN can be done using 12x less computations on this data set. Further, the computations consist of simple additions, which are energetically cheaper than the multiply-accumulate operations used in conventional ANNs. Future work concerns the measurement of the energy consumption (including the cost of memory transfer due to keeping neuron states).

To close the remaining accuracy gap, extensions of this work might consider training on the DVS events directly. This would create a purely asynchronous setting and potentially enable the model to accurately process streams with highly variable event rates. Regardless of the training method, detrimental inhomogeneities in the DVS input could potentially be removed by low-pass filtering or other preprocessing with smoothing effect.

The present work is a step towards efficient inference in mobile and embedded systems requiring low latency and computation cost, which will in particular profit from development of asynchronous event-based computing hardware. 

\begin{table}[t]
\caption{Classification accuracy of the original ANN and the converted SNN.}
\centering
\resizebox{0.4\textwidth}{!}{%
\begin{tabular}{ l c c }
         &  Accuracy  & Operations \\
Model (input type) & [\%] & [MOps] \\ 
\hline
\vspace{-0.2cm} \\
ANN (analog)& 88.25 & 7.85 \\
SNN (analog) & 88.12 & 1.15 \\
SNN (Poisson) & 86.77 & 3.06 \\
SNN (DVS) & 85.19 & 0.66 \\
\end{tabular}}
\label{tab:accuracy}
\vspace*{-1em}
\end{table}





\section*{ACKNOWLEDGMENT}
This work was funded by an SNSF project Ambizione under grant agreement PZOOP2\_168183.


\begin{small}
\bibliographystyle{IEEEtran} 
\bibliography{IEEEabrv,bibliography,references}
\end{small}

\end{document}